\documentclass[conference]{IEEEtran}
\usepackage{cite}
\usepackage{amsmath,amssymb,amsfonts}
\usepackage{algorithmic}
\usepackage{graphicx}
\usepackage{textcomp}
\usepackage{xcolor}
\def\BibTeX{{\rm B\kern-.05em{\sc i\kern-.025em b}\kern-.08em
    T\kern-.1667em\lower.7ex\hbox{E}\kern-.125emX}}

\usepackage{hyperref}
\usepackage{tikz}
\usetikzlibrary{er,arrows,positioning,fit}
\tikzset{
    >=stealth',
    punkt/.style={
           rectangle,
           rounded corners,
           draw=black, very thick,
           text width=6.5em,
           minimum height=2em,
           text centered},
    pil/.style={
           ->,
           thick,
           shorten <=2pt,
           shorten >=2pt,}
}
\newcommand{\revised}[1]{#1}
\newcommand{\intcomment}[1]{}

\begin{document}

\title{Neuromorphic Programming: Emerging Directions for Brain-Inspired Hardware}
% \title{Emerging Directions for Programming Brain-Inspired Computing Systems}
% \title{Neuromorphic Programming: Emerging Directions for Brain-Inspired Hardware}
% \title{Neuromorphic Programming: Harnessing Large-Scale Brain-Inspired Hardware}
% \title{Programming paradigms for brain-inspired hardware}
% \title{New Directions in Neuromorphic Programming}
% I like this one because it fits in one line - and brain-inspired hardware may make it more approachable also for people form ML/AI (beyond neuromorphic computing)
% Old title: Concepts and paradigms for neuromorphic programming
% Other suggestions:
% - Programming paradigms for neuromorphic [computing/hardware/systems]
% - Future directions in neuromorphic programming
% - Frontiers in neuromorphic programming
% - Advancing neuromorphic programming
% - Toward principled programming of neuromorphic hardware
% (consider replacing neuromorphic <> brain-inspired)

\author{\IEEEauthorblockN{Steven Abreu}
\IEEEauthorblockA{\textit{CogniGron Center \& Bernoulli Institute} \\
\textit{University of Groningen}\\
Groningen, Netherlands \\
s.abreu@rug.nl}
\and
\IEEEauthorblockN{Jens E. Pedersen}
\IEEEauthorblockA{\textit{Computational Science and Technology} \\
\textit{KTH Royal Institute of Technology}\\
Stockholm, Sweden \\
jeped@kth.se}
}
% \author{\IEEEauthorblockN{Authors hidden for peer review}}

\maketitle

\begin{abstract}
% motivation - programming crucial, current methods lacking, future outlook promising
The value of brain-inspired neuromorphic computers critically depends on our ability to program them for relevant tasks. Currently, neuromorphic hardware often relies on machine learning methods adapted from deep learning.
However, neuromorphic computers have potential far beyond deep learning if we can only harness their energy efficiency and full computational power. 
% claim: paradigm shift necessary
Neuromorphic programming will necessarily be different from conventional programming, requiring a paradigm shift in how we think about programming.
This paper presents a conceptual analysis of programming within the context of neuromorphic computing, challenging conventional paradigms and proposing a framework that aligns more closely with the physical intricacies of these systems. 
%%%%%
% We contribute a conceptual analysis that rethinks the role and definition of programming for neuromorphic systems and a detailed examination of promising, yet underutilized, programming paradigms. By sketching out a few promising, yet underutilized, techniques that leverage the unique properties of neuromorphic computers, this work aims to liberate researchers from the constraints of current approaches and encourage the exploration of new innovative strategies.
Our analysis revolves around five characteristics that are fundamental to neuromorphic programming and provides a basis for comparison to contemporary programming methods and languages.
% We contribute a conceptual analysis that rethinks the role and definition of programming for neuromorphic systems by defining five axes that capture the differences across the numerous types of programming models.
% By harnessing the successes of past approaches, we draw on promising, yet underutilized, techniques towards increasingly rich abstractions and improved productivity when programming neuromorphic devices.
% to leverage the unique properties of neuromorphic computers and compete with existing, conventional hardware platforms.
\revised{By studying past approaches, we contribute a framework that advocates for underutilized techniques and calls for richer abstractions to effectively instrument the new hardware class.}

\end{abstract}


% THREADS
% - DIVERSITY of approaches: hardware and programming paradigms (many different tools)
% - beyond digital: mixed-signal and analog
% - computing with physics
% - widen the _sense_ of 'programming'

\intcomment{A critique and recommendation for improving the manuscript is that the abstract and introduction make it a bit unclear whether this is a survey/perspective or whether it is providing a theoretical contribution. As it does a bit of both, the overarching message gets a little lost at times. For instance, the section IV discussion following the section III framework can come across as if there's still something more tangible and profound to come. And there is - but just not the codified programming paradigm presented here. And so my recommendation is to not oversell providing an answer in this work, but to rather emphasize the perspective and charge to the community.
TODO1: modify abstract to make clear that we only list things and point to things..
TODO2: read through everything once and improve the red thread throughout the paper
}

\section{Introduction}

Computing technology is steering toward an impasse, with Dennard scaling ending and Moore's law slowing down \cite{Waldrop2016}. The impasse gives rise to innovation opportunities for specialized hardware in computer architecture \cite{HennessyPatterson2019} as well as in software \cite{Edwards2021}.
This `Golden Age' of innovation has led many researchers to investigate neuromorphic computers. Taking inspiration from how the brain computes has a rich history \cite{Neumann1958} and the recent success of deep learning has demonstrated the power of neural information processing \cite{LeCunEtAl2015}. 
% The development of event-based sensors \cite{GallegoEtAl2022Event}, large-scale neuromorphic processors \cite{Schuman2017}, and brain-computer interfaces \cite{FlesherEtAl2021brain} indicates that neuromorphic computing is on the rise and is expected to play an important role in the future of computing \cite{Mead_2023}.
The development of event-based sensors, large-scale neuromorphic processors, and brain-computer interfaces indicate that neuromorphic computing is on the rise and expected to play an important role in the future of computing \cite{Mead_2023}.

Neuromorphic computers take inspiration from the brain, both in the way information is processed and in the fact that the physical dynamics of the underlying substrate are exploited for computation \cite{Jaeger2023}. 
A neuromorphic computer is composed of neurons and synapses which model biological neural networks. 
The hardware can either directly implement these biological neural network models physically \cite{IndiveriEtAl2011} or use them as inspiration and guidelines to design a more general, flexible, and configurable architecture \cite{Frenkel2021}. Key features of these biological networks include sparse connectivity, event-based communication, low-precision activations, and a focus on temporal processing.
%Although some artificial neural networks (ANNs) can be considered neuromorphic, the present paper focuses on spiking neural networks (SNNs) because they elegantly combine continuous-time analog dynamics and sparse event-based computing, making them an interesting topic for research on novel programming methods. 

The departure from fundamental assumptions in classical computing brings an urgent need for new theories to describe the computations in novel neuromorphic hardware, along with new theories and methods of programming that can make these devices useful.
The former has been outlined in recent work \cite{Jaeger2021,Jaeger2023} whereas the latter is constrained to an as-yet limited set of ``spiking neuromorphic algorithms'' \cite{SchumanEtAl2022Opportunities,Aimone2019}. Schuman \textit{et al.} \cite{SchumanEtAl2022Opportunities} argue that progress on neuromorphic programming requires a paradigm shift in how to think about programming.
Current programming methods are adapted to \textit{clocked digital hardware}, but with the forthcoming diversity of computer hardware and architectures \cite{HennessyPatterson2019} it is time to widen the set of hardware systems supported by our programming models.

\begin{figure}[h]
    \centering
    \includegraphics[width=0.95\columnwidth]{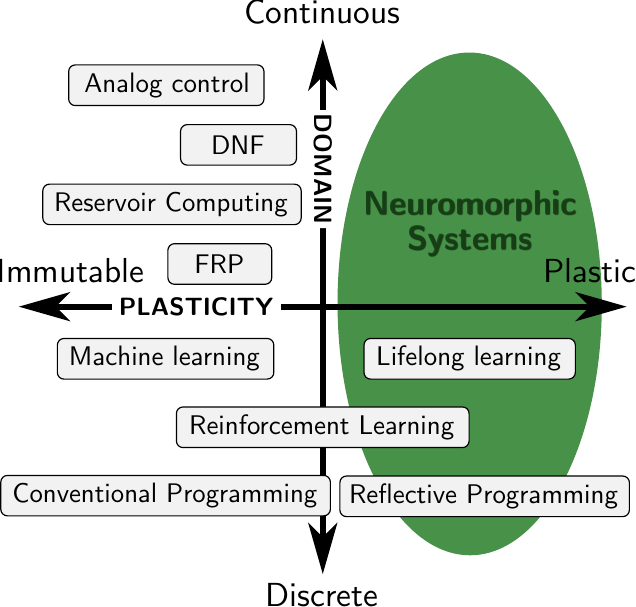}
    \caption{
    Programming and computational models plotted against their operational domain (continuous vs. analog) and the malleability of the computation during execution (immutable vs. plastic).
    The class of neuromorphic systems is shown as a green ellipse, spanning both the continuous and discrete domains.
    There is a lack of models for continuous and plastic systems.
    }
    \label{fig:hero}
\end{figure}

However, we need not start from scratch when designing methods and paradigms for neuromorphic programming. A lot of work has already been done that we can draw inspiration from \cite{Frenkel2021}, and new hardware is known to rejuvenate unconventional research ideas, as demonstrated by GPUs reviving research on neural networks trained with gradient descent \cite{Hooker2020,LeCunEtAl2015}. With the emergence of cutting-edge neuromorphic hardware, revisiting unconventional programming paradigms can uncover promising new directions \cite{BanatreEtAl2005}.

%%%%%%% Overview of the paper
In this work, we discuss the practical implications of programming neuromorphic systems and compare them with conventional methods in computer science.
We extend traditional concepts of programming to embrace the uniquely physical and adaptive properties of neuromorphic systems \revised{so as to} leverage these systems more effectively. 
Our contributions are twofold: first, we provide a detailed conceptual framework in Section \ref{s:theoretical-framework} that redefines programming in the context of neuromorphic hardware, challenging traditional paradigms and encouraging the adoption of novel approaches. Second, we identify and discuss advanced programming models in Section \ref{s:programming-paradigms} that are currently underutilized in the field, proposing new directions for their evolution and integration into mainstream computing practices. 

\section{Motivation} \label{s:motivation}

Neuromorphic hardware leverages the computational principles of the brain, which are vastly different from those exploited by conventional digital computers \cite{Neumann1958}.
These fundamental differences pose challenges for traditional programming abstractions, and it is clear that we cannot apply conventional theoretical computer science to understand the computational processes in neuromorphic systems \cite{Aimone_Parekh_2023}.
We highlight five fundamental differences underpinning this incompatibility.

\paragraph{Domain} Neurons are physical systems and operate in continuous time (see Figure \ref{fig:hero}).
The merits of analog computation in terms of energy efficiency and inherent parallelism are well-known \cite{Boahen2017,Sarpeshkar1998}. Classical symbolic computation, in contrast, is decoupled from real physical time and time is only \textit{simulated} through a discrete global clock signal. 
% But analog computing is more sensitive to device mismatch and noise which limits the computational depth (number of operations performed in series) \cite{Neumann1958}.
% They may also be susceptible to parameter drift, aging effects, and changes in temperature.
% %because signals are not restored at every timestep
% %because no two analog chips behave exactly the same

\paragraph{Plasticity} When programming digital computers, one may neglect the physical properties of the underlying hardware, as the underlying hardware does not change during execution. 
In neuromorphic computers, such hardware-agnostic programming is not generally possible, as these devices are by definition \emph{physical}.
Any programming model of neuromorphic systems must, therefore, include the malleability of the system, since the physical system and the computational model are one and the same.

\paragraph{Stochasticity} Unlike deterministically switching transistors, neural systems are \emph{stochastic}, which has lead to models of computation with probabilistic logic \cite{Neumann1956} and stochastic computing \cite{AlaghiHayes2013}, where information is represented in probability distributions.

\paragraph{Decentralization} The \emph{distributedness} of information representation and processing in neural networks stands in contrast to the localized information in binary transistor states and the sequential execution of elementary instructions in digital hardware. Such distributedness is leveraged in deep learning \cite{LeCunEtAl2015}, in dynamic neural fields \cite{Giese_1999} where neurons are considered independent independently evolving dynamical systems, and in hyperdimensional computing \cite{Kanerva2009} where high-dimensional random vectors are used for information representation and computation.

\paragraph{Unobservability} Neuromorphic hardware, especially analog and mixed-signal systems, often show limited observability, in that the system state can only be read out in parts. This is a key difficulty for plastic computations that change over time. It also relates to mismatches between platforms---a known challenge for analog devices that can also be observed between digital systems \cite{pedersen_neuromorphic_2023}.
% both for physical systems (e.g. analog chips), but also for plastic programs (they drift over time, and then we don't know what exact function is being implemented anymore). this is also related to mismatch, which is a known challenge for analog devices, but also across different digital systems (cite NIR). 

Figure \ref{fig:hero} fits programming and computational \revised{concepts} onto two axes: plasticity and domain.
It is immediately clear that the upper right quadrant is largely left unexplored.
To explain the vacuum, it is necessary to expand our concept of classical computation to include neuromorphic, physical systems.

\section{Theoretical framework}
\label{s:theoretical-framework}

The present section introduces a theoretical framework that abstracts ``computation'' to capture both classical and neuromorphic systems.
We begin by conceptualizing how neuromorphic computing fits within the broader landscape of computational models, highlighting areas that remain underexplored and elucidating why these gaps exist. 
% We aim to reconcile neuromorphic computing with classical computational concepts, focusing on how physical systems can be systematically integrated into current computational models.

\subsection{Computing with physical systems}

\begin{figure}[h]
    \centering
    \includegraphics[width=0.85\columnwidth]{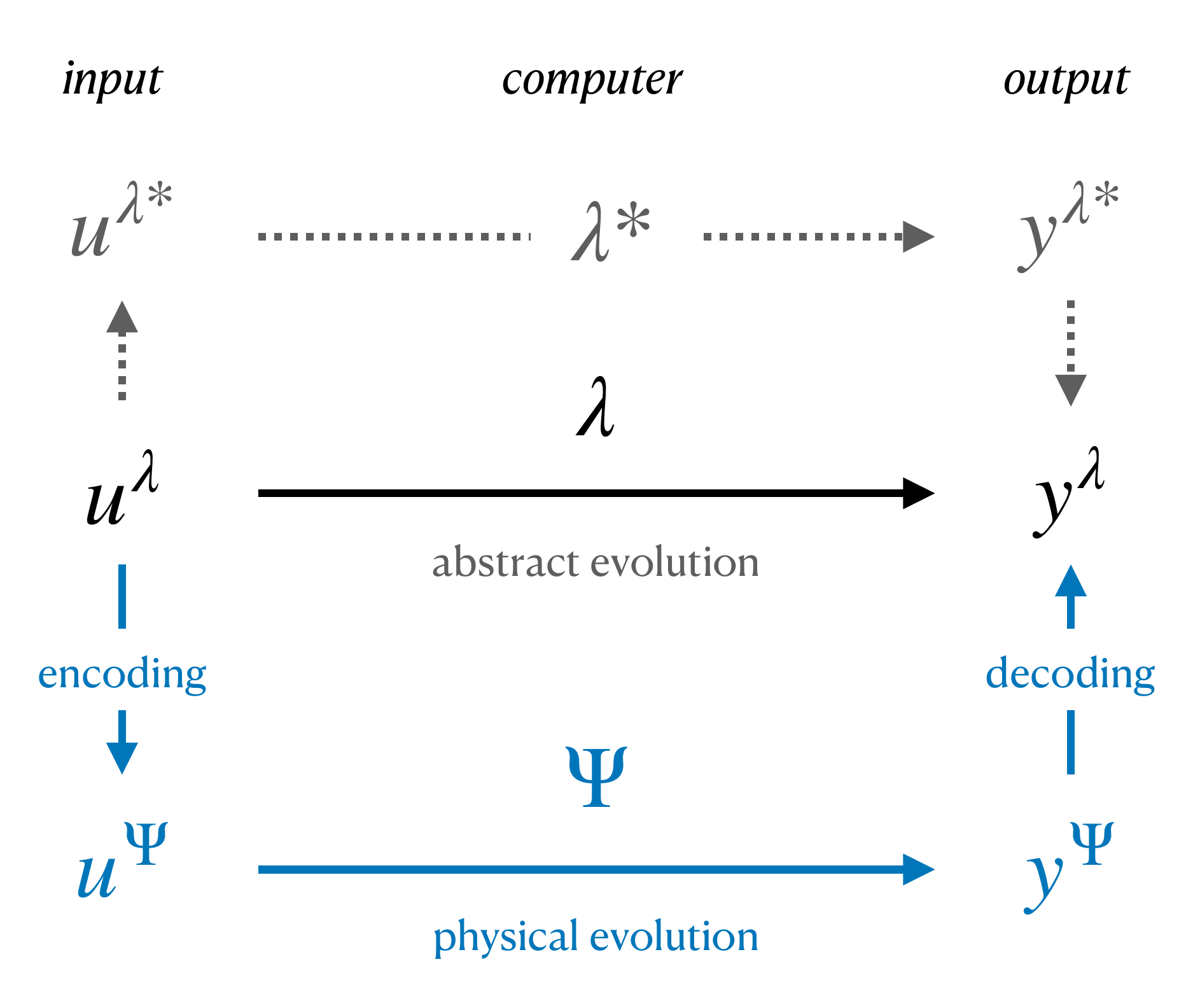}
    \caption{The relationship between an abstract computation $\lambda: u^\lambda \mapsto y^\lambda$ (middle), the same computation expressed in a different computational model $\lambda^*$ such as a higher-level programming language (top), and the physical computer $\Psi$ that realizes this computation (bottom), from \cite{HorsmanEtAl2014}.}
     % \textbf{Right} shows the process of computer programming, inspired by \cite{Gruenert2017}. The abstract intention lives in the informal space $\Omega$ of all possible computational tasks. This intention is formalized into a program $p \in \lambda$ that is expressed in some formal language, which is then implemented through some physical system $s \in \Psi$.
    \label{fig:commuting-diagram}
\end{figure}

Horsman \textit{et al.} \cite{HorsmanEtAl2014} provide a general framework for computation with arbitrary physical systems.
Therein, a computer is a physical machine $\Psi$ which can be stimulated by an input signal $u^\Psi$ and from which an output signal $y^\Psi$ can be read out. 
The computation $\lambda$ is specified by an abstract function from input $u^\lambda$ to output $y^\lambda$.
The machine $\Psi$ then implements the computation $\lambda$ if an encoding procedure $E$ and decoding procedure $D$ is known such that the machine $\Psi$ will produce $y^\Psi$ with $D(y^\Psi) \approx y^\lambda$ when stimulated with the input signal $E(u^\lambda)=u^\Psi$. 
This leads to the general form of the abstract computer model shown in Figure \ref{fig:commuting-diagram}: the physical machine $\Psi$ receives input $u^\Psi$ and produces output $y^\Psi$, thereby implementing the abstract computation $\lambda$ from input $u^\lambda$ to output $y^\lambda$. 

Any machine that is changing during program execution will, according to Horsman \textit{et al.} require either that the machine is insensitive to the change such that the effective input-output function stays unaffected, or that the computational model includes the change, so the computation stays correct.
Thus, to fit the framework of Horsman \textit{et al.}, neuromorphic computing first requires a model that captures their (plastic) operation. This can be expressed in the formalism of (non-autonomous) dynamical systems, logic-based systems or other formalisms \cite{Jaeger2021}.

% Although classical theories of computing are non-physical, all computations must ultimately be physically instantiated \cite{Deutsch1985}.
For digital computers, the abstract model of computation was developed first and only later physically realized. 
% Digital computing was first developed as an abstract model which was later physically realized.
In contrast, neuromorphic hardware does not rely on any universally accepted abstract model \cite{Schuman2017}. Abstract models of computation are co-developed with physical implementations \cite{abreu_photonics_2024}.
% From a physical perspective, the key difference between conventional computing and neuromorphic computing lies in the set of physical phenomena that are harnessed for computation. While digital computing only uses bi-stable switching dynamics, neuromorphic computers use stochasticity, real-valued states in continuous time, and more \cite{Jaeger2021}.

% \subsection{Computations and Programs}

%{\color{red}
%would be good to also clarify the relation between the (closely related, apparently) concepts of "program" (according to your view) and "algorithm" (according to the digital conception as "algorithm = Turing machine" for the theoretical view, "algorithm = program written in a programming language" for the practical view).
%}
\subsection{Computer programs}

While a computation $\mathcal{C}$ specifies \emph{what} is being computed, a program $\mathcal{P}$ specifies \emph{how} the computation is implemented. Many different programs $\mathcal{P}_1,\ldots,\mathcal{P}_n$ may implement the same computation $\mathcal{C}$. 
Note that the concept of a `program' herein includes algorithms as Turing machines as well as learning algorithms and interactive programs---although the latter two cannot be implemented by Turing machines \cite{Valiant2013,Wegner1997}.

% In classical computing, a function on natural numbers is implemented by a program which can be represented by a Turing machine.
% %This can also be expressed as a function on symbols strings (words), by encoding these symbol strings into numbers. 
% In neuromorphic computing, functions that operate on (real-valued) time series are computed. The computation is implemented by a program represented as a neural network, typically with designated input and output neurons.

A computation $\mathcal{C}$ is described by a formal specification that formalizes the intention of the computation (Figure \ref{fig:programming-process}). The specification of a computation is expressed in some mathematical formalism. In digital computing, this can be done using formalisms from logic. In analog computing, there are various formalisms that describe the computation, for example qualitative geometrical constructs like attractors and bifurcations \cite{Jaeger2021}.
%A specification is said to be complete if exactly one function satisfies the specification, and incomplete if more than one function satisfy the specification. 

A program $\mathcal{P}$ is described in another formalism. In digital computing, programs are expressed in some programming language. In analog computing, one typically uses differential equations to describe the program. When programs interact with another, one also speaks of each individual program as a \emph{process} and the ensemble of all processes as the program, whose behavior emerges from the interaction of the processes.

Operationally, a program is defined by the data flow and control flow.
% data flow and control flow
The data flow specifies how signals that carry computationally relevant information are propagated through the machine. 
The control flow specifies what operations or transformations are done on these signals. 
% examples
For example, in a field-programmable gate array (FPGA) the data flow is configured through its routing elements while the control flow is defined by the function implemented in each logic block. 
% In a CPU, data flows between registers and memory according to the program's instructions, while the control flow is defined by its logic instructions.
In a neuromorphic chip, the data flow is defined by the connectivity of the neural network while the control flow is defined by the synapse and neuron models, learning rules, synaptic weights, time constants, thresholds, and more.
%For example, each logic block in an FPGA can be configured to implement some function on its incoming data. In a CPU, the control flow is defined by its instructions that do not specify data transfer only. In a neuromorphic chip, the control flow is defined by the synapse and neuron models, local learning rules, weights of the network, and additional parameters like gains and time constants.

%The programming language provides a number of elementary operations which can be composed. 

%The classical definition of digital computation does not apply to neuromorphic computers. Digital computation was first proposed as an abstract model which was only later physically realized. As such, the theory is inherently unphysical. 
%A computation $\mathcal{C}$ is a function from input signals $u$ to output signals $y$ which can be implemented by a program $\mathcal{P}$. The program $\mathcal{P}$ defines an abstract evolution of 

\subsection{Programming process}

\begin{figure}[h]
    \centering
    \includegraphics[width=0.99\columnwidth]{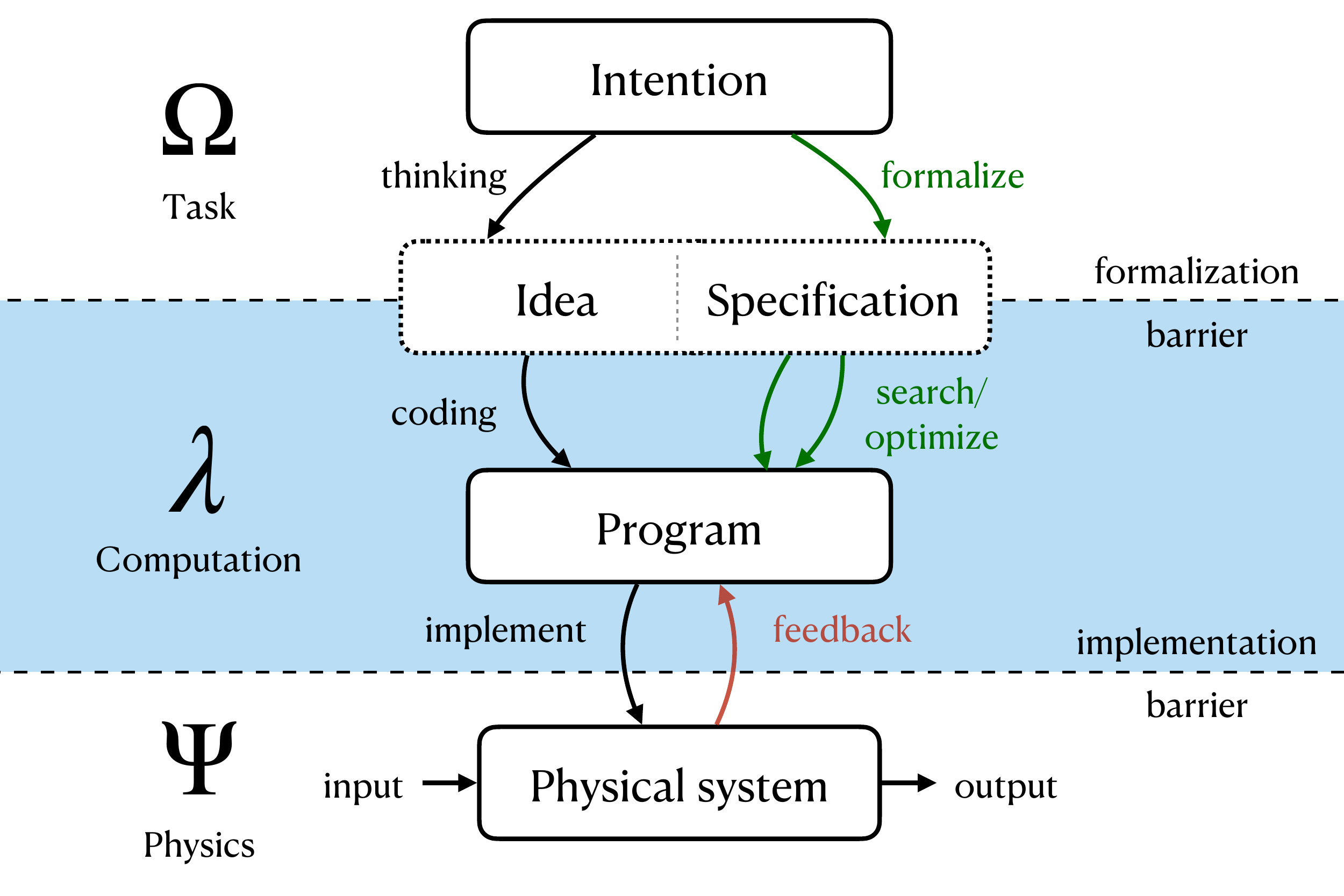}
    \caption{The process of computer programming, inspired by \cite{Gruenert2017}. The abstract intention lives in the informal space $\Omega$ of all possible computational tasks. This intention is formalized into a program $p \in \lambda$ that is expressed in some formal language, which is then implemented through some physical system $s \in \Psi$.}
    \label{fig:programming-process}
\end{figure}

Programming, in the context of the theoretical framework above, is the process of designing \textit{how} a certain computation should be implemented, illustrated in
Figure \ref{fig:programming-process}. Programming begins with some informal intention of what computation to implement. This intention can be formalized into a specification (right path in Figure \ref{fig:programming-process}), or the programmer may directly come up with an informal idea for a program that implements the intended computation, and then code this idea in some formal language (left path in Figure \ref{fig:programming-process}). This program is then communicated to the physical computer through a pre-defined programming interface. Finally, the system executing this program can be controlled or instructed to remain within the program's specification.
% program need not be fully known/formalized. Can be just an incomplete idea. The idea can be completed by blanks which are learned/trained/evolved/searched.

% From this diagram, programming is the act of coming up with a program that satisfies some intention or specification, with no regard for how that program is found. This is a rather general notion of programming which goes beyond the notion that is typically associated with the term. 
% % put definition of programming from Haridi textbook?
% In common parlance, programming refers to the process in which a human programmer has an idea for how to solve a problem and codes their solution in some high-level programming language (left path on Figure \ref{fig:programming-process}, top). 
%
Programming need not be done by a human programmer. As shown by the green arrows in the diagram, programming can also be automated: a computer program can take in a formal specification of a program to then create a program that satisfies these specifications. 
In \textit{program synthesis}, the specification is given in some algebra or formal model and a search algorithm then finds a program that meets this specification. 
In \textit{supervised machine learning}, the specification is given by a dataset of input-output examples and an error function, which is then minimized for some machine learning model until it meets a given error threshold.

\subsection{Languages and Paradigms}
\label{ss:languages-paradigms}

Conventionally, programming amounts to writing source code, \emph{coding}, in some formal language. Here, `programming language' is used in an unconventionally wide sense to include any formal language that can be communicated to a physical system. 
This includes programming languages like Python but also extends to other formalisms like differential equations describing dynamical systems, or block diagrams describing signal processing systems. 
In any case, the `programming language' must be compatible with the elementary instructions that the computer's programming interface provides. Given this compatibility, the programmer is free to explore the space of all possible programs.
Work on elementary instruction sets for non-digital computers goes back at least to the 1940s and continues to the present day \cite{Shannon1941,Hasler2020a} but there is still no universally accepted model \cite{Jaeger2021}. 
Consequently, it is not clear what a neuromorphic programming language may look like \cite{MichelEtAl2006}; will it require new syntax such as visual representations, or will a program be represented by a string of symbols in some formal language?

Since the goal is to improve the way we think about programming neuromorphic computers in general, Floyd \cite{Floyd1979} argued that it is more effective to turn to \emph{programming paradigms} rather than programming languages. 
A programming paradigm is an approach to programming ``based on a mathematical theory or a coherent set of principles'' \cite{Roy2009} and a programming language implements one or more programming paradigms.
Centering the discussion on programming paradigms shifts the focus away from syntactical issues to the way programs are conceived and designed.

Beyond languages and paradigms, there is a set of well-developed tools for computer programming without which most modern software systems would not exist. We mention the integrated development environment (IDE), the keyboard-and-mouse interface, version control systems, agile development only in passing, as a more detailed treatment of these is outside the scope of this paper. %However, we stress that such tools will be necessary for neuromorphic programming to develop into a mature discipline.

\section{Programming Paradigms}
\label{s:programming-paradigms}

In the present section, existing programming paradigms are explored and then related to Neuromorphic Programming in Section \ref{s:neuromorphic-programming}.
Conventional computer programming paradigms can be differentiated by the program's representation. 
One commonly distinguishes \emph{imperative programming}, where a program consists of a sequence of instructions, from \emph{declarative programming}, where a program consists of a sequence of logical assignments. 
We use the term \emph{decentralized programming} for parallel, concurrent, and distributed programming, spanning both imperative and declarative styles.
% Intuitively, imperative programs describe \emph{how} a program is executed while declarative programs describe \emph{what} the program should do.
We distinguish these paradigms from \emph{automated programming}, in which the human programmer passes a (potentially incomplete) specification into an optimization or learning algorithm that creates the final program.
% In reality, there is no clear delineation between manual and automated programming, as even the compilation from a high-level programming language into executable machine code can be seen as a form of automation. 
Finally, we also give a brief overview of \emph{non-digital programming} methods that are applicable to analog and physical computers.

\subsection{Imperative programming}
The most common way of writing sequential, instruction-based programs uses the \emph{imperative} paradigm, as implemented in C. 
Imperative programming was augmented with objects, which can contain instructions as well as data, to yield the \emph{object-oriented} paradigm, as implemented in C++ or Java. Imperative programming is the dominant paradigm for large-scale software systems in conventional computing. 

\subsection{Declarative}
Instead of describing the control flow of a program in imperative programming, declarative programs describe the logic of the program. A declarative program describes \emph{what} the program does rather than \emph{how} it does it. Declarative programming is done in database query languages like SQL, functional programming languages like Haskell, or logic programming languages like Prolog.
In \emph{dataflow programming}, a program is modeled as a graph of data flowing between operations. 
% \textbf{Spatial programming} can be used to program reconfigurable hardware into dataflow engines \cite{BeckerEtAl2016}.
A functional cousin of dataflows is the \emph{functional reactive programming} (FRP) that couples hybrid systems of \textit{behaviors} and \textit{events} with continuous-time flow \cite{Wan_Hudak_2000}.
Dataflow programming is well-suited for neuromorphic computers, where data flows between neurons.
%This allows for \emph{massively} parallel computation, similar to cellular automata.

While classical programs are deterministic, the execution of a \textit{probabilistic program} depends on random numbers, for example by calling a (pseudo) random number generator. Such a program can be viewed as sampling from a probability distribution. In \emph{probabilistic programming}, the program itself is considered a distribution, and the programmer can analyze this distribution and condition the distribution on observations \cite{GordonEtAl2014}. Indeed, the goal of probabilistic programming is not simply the execution of a program, but also the analysis thereof. 
% By expressing a statistical model as a probabilistic program, statistical inference on such a model can be done automatically by the compiler through general-purpose inference schemes. 
% Probabilistic programming has been used for state-of-the-art generative vision models with very compact programs of only about 50 lines \cite{KulkarniEtAl2015}.
% \textit{Other than by influences of dataflow programming, neuromorphic computing typically does not leverage declarative of functional programmin methods.}

\subsection{Decentralized programming}
A single processor core may use time-sharing to allow multiple programs to run concurrently. In so-called \emph{concurrent programming}, the lifetime of multiple computing processes overlap and may interact with another \cite{Milner1993}. Concurrency introduces issues of synchronization, such as deadlocks and race conditions.
With the advent of multicore microprocessors came the need to use resources on \emph{different} cores simultaneously. This led to the development of \emph{parallel programming} techniques, in which multiple processes are carried out simultaneously on different cores. % \cite{PachecoMalensek2022Introduction}. 
% Functional programming makes reasoning about programs easier and simplifies parallel programming \cite{Backus1978}.
% Neuromorphic computers are inherently parallel, even \emph{massively parallel}, and concepts from parallel programming are often imported when dealing with neuromorphic systems.
%Amdahl's law \cite{-Amdahl1967} states that the speedup of parallelization is limited by the ratio of the program that is inherently sequential. 
%
%Petri nets are a general, graph-based formalism which can model many sorts event-driven, concurrent processes, not only parallel unclocked computer programs but also real-world systems like production processes in factories. 
\emph{Distributed programming} deals with programs that are executed on multiple networked computers, which interact to achieve a common goal.
The first approach to formalizing this appeared by Milner et al. as $\pi$-calculus \cite{Milner_1992}.
Methods from distributed programming are used for multi-chip systems, and frequently employed in training large-scale neural networks. 
% \textit{Concepts from the above-mentioned decentralized programming methods are frequently used in programming neuromorphic systems.}

\subsection{Automated programming}
In \emph{meta-programming}, it is possible for a program to write or modify programs, by simply treating the program as data. 
In \emph{reflective programming}, a program modifies its own behavior whereas in \emph{automatic programming}, a program generates another program. 
If a formal specification of the desired program is given, \emph{program synthesis} can be used to generate a program that provably satisfies this specification \cite{GulwaniEtAl2017}.
If exact adherence to a formal specification is not required, but only the satisfaction of given constraints, \emph{constraint programming} may be used \cite{RossiEtAl2008Constraint}.
If an incomplete specification is available, such as input-output examples, then \emph{inductive programming} can be used to generate a suitable candidate program \cite{GulwaniEtAl2015Inductive}.
%The idea of inductive programming may be even further extended to natural language programming (NLP, see Figure \ref{fig:programming-diagram}) where a computer program generates a desired program from an input given in natural language. 
% An inductive programming approach coupled with probabilistic programs has been proposed as a model for human-level concept learning \cite{LakeEtAl2015}.
% Recently, deep learning (see below) has been used for inductive programming, under the name of neural program synthesis \cite{Kant2018}. 
%end-user programming
% As already mentioned in Section \ref{ss:programming}, it is possible to instruct an interactive program and direct it to implement a desired computation. \textbf{End-user programming} allows users to obtain programs from a small set of examples, like the flashfill feature in spreadsheet programs, which infers a formula from table manipulations done by the user \cite{Schmid2018,GulwaniEtAl2015Inductive}. 

%%%%%%%%%%%%%%%%%%%% Machine learning, deep learning, differentiable, clarify connection to optimization, probabilistic
In classical programming, a human programmer defines the program that specifies how input data is processed. \emph{Machine learning} constructs programs that learn from the input data, in ways that may not have been anticipated by any human. Machine learning has deep roots in probability theory and overlaps significantly with probabilistic programming \cite{Murphy2012Machine}.
%For the present context, supervised machine learning and reinforcement learning are particularly relevant. 
In supervised machine learning, a mapping from inputs to outputs is learned from a set of examples. 
In reinforcement learning, a policy of how to act in some environment is learned from rewards and punishments. 
Both the learned mapping in supervised learning and the learned policy in reinforcement learning can be used as programs. This makes machine learning a paradigm for automated programming. % \cite{CholletAllaire2018Deep}.
\emph{Deep learning} uses multi-layered artificial neural networks (ANNs) for machine learning. The connectivity in ANNs is usually fixed, and the weights are learned, typically in a supervised fashion using gradient descent, to minimize the error on given input-output examples.
The techniques of deep learning have also been adapted for spiking neural networks \cite{Eshraghian2023training}.
In \emph{differentiable programming}, programs are written in a way that they are fully differentiable with respect to some loss function, thereby allowing the use of gradient-based optimization methods to find better-performing programs. Deep learning is a special case of this, where programs are artificial neural networks that are differentiated using backpropagation. % \cite{Fong_2019}. 
% Differentiable programming has been employed to merge deep learning with physics engines in robotics \cite{DegraveEtAl2019}, it has been applied to scientific computing \cite{InnesEtAl2019}, and even towards a fully differentiable Neural Turing Machine \cite{GravesEtAl2014}.

%%%%% Optimization
% Machine learning relies heavily on tools from optimization theory. In pure optimization, the minimization of some cost function $J$ is a goal in itself. 
In machine learning, a core goal is good generalization to unseen examples. 
% This is expressed as some performance measure $P$ which is intractable \cite{Wolpert_Macready_1997} and therefore one minimizes some cost function $J$ which will in turn also increase the performance measure $P$.
If generalization is not needed, one may use \emph{optimization} itself as a programming paradigm in which the result of the optimization is the desired program or the optimization process itself.
% https://cedar.buffalo.edu/~srihari/CSE676/8.1%20LearningVsOptimizn.pdf
For example, \emph{evolutionary programming} uses population-based evolutionary optimization algorithms to find programs by encoding the program's specification in a fitness function that is optimized. % To find a program that solves some problem one defines a fitness function that is maximized by a program that solves this problem.
Evolutionary algorithms have been used to generate rules for a cellular automaton to solve computational problems that are difficult to solve by manually designing a learning rule \cite{MitchellEtAl1994}. 
%
%
% \textit{Automated programming methods from deep learning, reservoir computing, and evolutionary optimization are popular approaches for programming neuromorphic systems \cite{SchumanEtAl2022Opportunities}.}
% Evolutionary optimization is a popular approach for neuromorphic systems \cite{SchumanEtAl2020Evolutionary}.

% \paragraph{Cellular programming}
% As mentioned previously, cellular automata (CA) are a standard model of massively parallel computation. A CA is programmed by choosing its update rule and the program is executed on some initial configuration of the CA's lattice.
% Inspired by CAs, cellular architectures of neuromorphic devices have been proposed \cite{KhacefEtAl2020Brain,Schuman2017}.
% For over two decades, \textbf{amorphous computing} has been developing programming techniques inspired by the cellular cooperation in biological organisms \cite{Stark2013Amorphous}. An amorphous computer is a system of irregularly placed, asynchronous, locally interacting computing elements that are possibly faulty, sensitive to the environment, and may generate actions \cite{AbelsonEtAl2000,Coore2005Introduction}. This line of research brought space-time programming \cite{BealViroli2015} as a way of programming to control large networks of spatially embedded computers. 
% Although not directly focused on neuromorphic computing, amorphous programming methods can provide a good starting point for robust programming methods in cellular architectures.

\subsection{Non-digital programming}
% \paragraph{Analog programming}
%In neuromorphic computers, and other unconventional computers \cite{Adamatzky2018Unconventional}, the computation is directly implemented in the physics which builds a close interplay between physics and computing that is not present in digital computing and which current cannot be fully exploited for a lack of theory \cite{Jaeger2021}. Nevertheless, approaches exist which can harness the dynamics of a physical system for computation. 
Building on evolutionary optimization, \emph{evolution in materio} \cite{MillerDowning2002} was proposed to harness material properties for computation. Natural evolution excels in exploiting the physical properties of materials, and artificial evolution aims to emulate this ability.
% Evolution has been applied widely in unconventional computing \cite{Adamatzky2018Unconventional}, for example with a disordered dopant-atom network for digit classification \cite{ChenEtAl2020}.
\emph{Physical reservoir computing} can be used to harness the dynamics of physical systems for computation by modeling the physical system as a high-dimensional reservoir for which a linear readout is trained \cite{TanakaEtAl2019}.
It is also possible to create a \emph{surrogate model} of the physical device, then optimize the surrogate model in simulation with deep learning methods and transfer the optimized model back to the device \cite{wright_deep_2022}.
%
%%%%%%% Analog
Although analog computers have been around for at least as long as their digital counterparts, \emph{analog programming} methods are not at the same level of maturity as those for digital computers. Ulmann \cite{Ulmann2020} argues that the development of reconfigurable analog computers will advance the state of analog computer programming, and efforts to develop such hardware already exist \cite{Hasler2020}.
Currently, analog programming often draws on methods from control engineering, signal processing and cybernetics.
% While digital computing was originally formulated as computing functions on the integers \cite{Turing1937}, \emph{signal processing} can be seen as computing functions on temporal signals. 
For analog neuromorphic computers, \emph{signal processing} provides a rich framework for computing with temporal signals \cite{DonatiEtAl2018Processing}.
Moreover, \emph{control theory} has developed a rich repertoire of methods to drive a dynamical system into a mode of operation that is robust, stable, and implements some desired dynamics \cite{Kirk_1970}. These methods can be used to keep analog computers within a desired regime of operation to implement a desired computation.
However, the expressiveness of behaviors in control theory lags far behind that of digital programming languages.
% It can be expected that analog computers can benefit from cross-fertilization between computer science and control theory \cite{MichelEtAl2006}.
% A promising direction is data-driven control where a model of the system to be controlled is learned from experimental data using machine learning techniques \cite{BruntonKutz2019Data}.
% Historically rooted in ideas from cybernetics and ultrastable systems \cite{Ashby1960Design}, 
The field of \emph{autonomic computing} aims to design systems that adapt to stay within a high-level description of desired behavior \cite{ParasharHariri2005}. The field takes inspiration from the autonomic nervous system, which can stay within a stable `dynamic equilibrium' without global top-down control.
% Computational models, and therefore programming methods, must ultimately be based in physics and resulting hardware constraints \cite{Stepney2012a}. 
% \textit{Methods from signal processing, control theory and physical reservoir computing are actively being used for programming neuromorphic systems.}

\section{Neuromorphic Programming}
\label{s:neuromorphic-programming}
%DC has had decades to build abstraction and compilation hierarchies to facilitate the programming of increasingly complex computer systems.

The preceding section already alluded to some methods that have been adopted for neuromorphic hardware, such as reservoir computing, deep learning, and evolutionary optimization. In the present section, we give a more in-depth review of existing programming approaches for neuromorphic hardware.

\paragraph{Neuromorphic co-design}
% manual design -> starting point
% gives an upper bound for the flexibility one can have
As neuromorphic computers exploit physical phenomena of their underlying hardware, manually designed neuromorphic programs will necessarily be close to physics. Therefore, although not strictly a `programming' paradigm, it is instructive to consider \textit{neuromorphic co-design} as a paradigm for designing neuromorphic systems.
The field is rooted in the original vision of neuromorphic computing \cite{Mead1990} and designs application-specific and reconfigurable mixed-signal neuromorphic chips in sub-threshold CMOS technology. 
This approach uses tools from signal processing and computational neuroscience to implement a desired behavior in networks of silicon neurons \cite{IndiveriEtAl2011}. 
% Similar to analog computing, the field may benefit from a set of computational primitives to simplify the design of neuromorphic systems.

\paragraph{Machine learning methods}
Given the success of deep learning in applying machine learning techniques to neural networks, this is a natural start\revised{ing} point for neuromorphic computers. 
% Given the success of deep learning, learning is a natural paradigm for neuromorphic computers. 
However, it is unrealistic to expect deep learning methods to work for SNNs as well as they do for ANNs since these methods were optimized for ANNs \cite{DaviesEtAl2021Advancing}. 
%
% To compile a neural network into hardware, it is necessary to first design the neural network's architecture. Deep learning has accumulated a plethora of well-performing network architectures for ANNs 
\emph{ANN-to-SNN conversion} is possible, but typically not optimal because the resulting SNNs do not leverage the computational power of spiking neurons. Instead, they limit the richer dynamics of SNNs to the less expressive domain of ANNs \cite{SchumanEtAl2022Opportunities}.
Offline training methods like \emph{backpropagation}, the workhorse of deep learning, can be implemented directly in SNNs using surrogate gradients \cite{Eshraghian2023training}.
% Simplifications of the backpropagation algorithm such as the random backpropagation algorithm \cite{BaldiEtAl2018Learning} were also demonstrated in neuromorphic systems \cite{NeftciEtAl2017}. 
% 
Given a neural network, it is necessary to communicate this network to the hardware. 
\emph{Neuromorphic compilation} \cite{ZhangEtAl2020} was proposed as a general framework to (approximately) compile neural networks into different hardware systems, automatically adapting to physical constraints. 
% Such compilation can be done statically to exactly implement the specified network architecture \cite{GruauEtAl1995neural,Siegelmann1994Neural}, or adaptively to further optimize the network after compilation \cite{BunelEtAl2016Adaptive}. In any case, it is important to consider the hardware constraints in this compilation \cite{JiEtAl2018Bridge}.
%
% Compilation and conversion are promising directions, though descriptions at the level of neural network architectures may not provide a high enough abstraction for implementing programs that realize arbitrary computations. 
\emph{Reservoir computing} can simplify the training of SNNs on hardware, but it still requires the reservoir states to be read out and stored on a digital computer, which may not be possible given limited observability of some devices or limited data storage. 
% Reservoir computing is a popular paradigm for neuromorphic computing, with dedicated frameworks for hardware implementation \cite{SchumanEtAl2022Opportunities,Michaelis2020}.

\paragraph{Online learning}
The machine learning methods presented above all operate offline and often off-device. Frequent re-training creates a large overhead, limiting the performance and applicability of neuromorphic computers. 
As a result, \emph{on-device learning} methods are an active topic of research \cite{BasuEtAl2018Low}. 
\emph{Plasticity} is a popular paradigm for on-device learning, where local learning rules are used to modify the connectivity (structural plasticity) and connection strengths (synaptic plasticity) of a SNN.
Parallels to emergent programming may be drawn here, as the resulting behavior of the SNN emerges from the interaction of local rules. It is not clear what local rules will yield a particular network-level behavior, but evolutionary search \cite{JordanEtAl2020} and meta-learning \cite{ConfavreuxEtAl2020} have been used to (re-)discover desirable plasticity rules.

\paragraph{Evolutionary methods}
A key advantage of evolutionary approaches is that they can jointly optimize the network's architecture and weights, thus simultaneously designing and training the network without requiring the network to be differentiable.
% Evolutionary approaches can find an SNN by randomly choosing an initial population of candidate SNNs, selecting the highest-performing candidates according to some performance metric, and then creating new candidates through recombining and mutating the selected candidates \cite{SchumanEtAl2020Evolutionary,SchliebsKasabov2013Evolving}. 
However, evolutionary approaches can be slower to converge than other training methods and the resulting architectures are not easily interpretable or reusable for different tasks \cite{SchumanEtAl2022Opportunities}.

\paragraph{Spiking neuromorphic algorithms}
With the increased availability of neuromorphic hardware, a number of handcrafted spiking neuromorphic algorithms (SNA) have been proposed. SNAs implement computations using temporal information processing with spikes, often to implement well-defined computations such as functions on sets of numbers \cite{VerziEtAl2018Computing}, functions on graphs \cite{HamiltonEtAl2019Spike}, solving constraint satisfaction problems or solving a steady-state partial differential equation using random walks \cite{SmithEtAl2020Solving}.
% SNAs are being actively developed and many application domains are yet to be explored \cite{Aimone2019}. This is perhaps the closest analogue to manual conventional programming, in which hardware primitives are used to build up complex computations.
% A number of handcrafted SNN algorithms have been proposed in recent years to solve well-deﬁned computational problems using spike-based temporal information processing. When implemented on neuromorphic architectures, these algorithms promise speed and efﬁciency gains by exploiting ﬁne-grain parallelism and event-based computation. Examples include computational primitives, such as sorting, max, min, and median operations [70], a wide range of graph algorithms [71]–[74], NP-complete/hard problems, such as constraint satisfaction [75], boolean satisﬁability [76], dynamic programming [77], and quadratic unconstrained binary optimization [78], [79], and novel Turing-complete computational frameworks, such as Stick [80] and SN P [81].

% primitives and synthesis
\paragraph{Neurocomputational primitives}
Various neurocomputational primitives have been proposed in the neuromorphic community. Such primitives can be useful for simple tasks and can be combined into complex neuromorphic systems \cite{BartolozziEtAl2022Embodied}.
% WTA & DNF
The winner-take-all (WTA) network is a common circuit motive in the neocortex that has been used extensively for neuromorphic systems \cite{DouglasMartin2007Recurrent}. The more general \emph{dynamic neural fields} (DNFs) are a modern framework for neural attractor networks \cite{Giese_1999}. 
% The stable states provided by attractor dynamics help with the intrinsic variability of analog neuromorphic circuits and have been shown to be a promising abstraction for neuromorphic programming \cite{DaviesEtAl2021Advancing}.
% Each DNF is a network of neurons that is, under some constraints, computationally equivalent to a \textbf{winner-take-all} (WTA) network \cite{Sandamirskaya2014Dynamic}. The WTA is a common circuit motive in the neocortex \cite{DouglasMartin2007Recurrent}.
%Computation with attractor networks has a rich history in neural computation \cite{Hopfield1982} and provides a simple way for ....... neurosymbolic integration.
% NSM
The \emph{neural state machine} (NSM) \cite{NeftciEtAl2013} also builds on WTA networks to implement finite state machines in SNNs.
% , and has been shown to run robustly on mixed-signal neuromorphic hardware.
% % sPLL
% The \textbf{spiking phase-locked loop} (sPLL) \cite{MastellaChicca2021} was designed for frequency detection as part of a neuromorphic tactile sensor.
% TDE
The \emph{temporal difference encoder} (TDE) \cite{GutierrezGalanEtAl2021Event} is a circuit primitive that has been used for motion estimation and obstacle avoidance.
% spiking model that was designed to compute the time difference between two consecutive input spikes. The number of output spikes and the time between them is inversely proportional to the time difference. This has been used for motion estimation and obstacle avoidance \cite{MildeEtAl2017Obstacle}.
%Delay/Temporal measurement circuits take inspiration from the insect brain, where motion is computed as the time to travel of a stimulus from one sensing element to the neighbour 180 . This type of computational primitive is useful for motion estimation and obstacle avoidance 88 
% for temporal encoding on event-based signals, which translates the time difference between two consecutive input events into a burst of output events. The number of output events along with the time between them encodes the temporal information.
%The TDE unit translates the time difference between two events into a burst of output spikes. Both the number of output spikes and the duration of the burst produced by the model directly reflect the temporal correlation of two input signals, and it is inversely proportional to the time difference
\emph{Neural oscillators} generate rhythmic activity that can be used for feature binding and motor coordination, for example as a central pattern generator \cite{KrauseEtAl2021Robust}.
% Further
Other primitives are scattered around the literature, and shared libraries of neurocomputational primitives are only starting to be assembled \cite{BartolozziEtAl2022Embodied}.
%
% \textbf{Neuromorphic synthesis} \cite{NeftciEtAl2013} may provide a systematic way of programming complex high-level behavior into neuromorphic chips. This was demonstrated for functions that can be described by finite state machines, but it may be promising to extend this work to a larger set of computational primitives for higher abstractions in neuromorphic programming.

\paragraph{Higher abstractions}
% NEF
The \emph{neural engineering framework} \cite{EliasmithAnderson2002Neural} raises the level of abstraction beyond networks of neurons---it allows dynamical systems to be automatically distilled into networks of spiking neurons using the Nengo programming environment \cite{BekolayEtAl2014}.
% VSA
The framework of \emph{vector symbolic architectures} (VSA) is suitable for neuromorphic systems, enabling knowledge representation and reasoning in high-dimensional spaces.
% Vector symbolic architectures (VSAs) offer a mathematical, connectionist framework that supports rich knowledge representations and reasoning in high-dimensional spaces. 
VSAs interfaced with deep networks and generalizations of the optimization and search algorithms described in this survey could provide a path to enabling fast, efficient, and scalable next-generation AI capabilities on neuromorphic hardware.
% Lava
\emph{Lava} is an open-source neuromorphic programming framework that includes libraries of neuromorphic algorithms for optimization, attractor networks, deep learning methods for SNNs, VSAs, and plans to include more paradigms.
% Fugu
\emph{Fugu} \cite{AimoneEtAl2019Composing} is a hardware-independent mechanism for composing SNAs. In Fugu, a program is specified as a computational graph, reminiscent of dataflow programming, where nodes represent SNAs and connections represent dataflow between the SNAs. 
\section{Future approaches to programming brain-inspired hardware}
\label{s:future}

%Programming is, by definition, a anthropocentric approach, and the abstra
% computing and physics gap
% Formalizing, making protocols, defining standards
For neuromorphic systems to scale to large heterogeneous computing systems, commonly agreed-upon computational models are required, similar to how programming abstractions catalyzed the development of the digital computer in the 1950s and 60s.
%As the field is moving toward generally applicable programming methods, it is crucial to clarify concepts and establish an efficient separation of concerns to allow effective cross-disciplinary collaboration and communication.
% 1. There is a gap
We have argued that it is difficult to program computers that harness their underlying physical dynamics for computation without a guiding theory that unites physics with computation.
But, despite the rich history of programming methods and languages, our analyses in Sections \ref{s:motivation} and \ref{s:theoretical-framework} show that the direct translation of classical methods to neuromorphic systems is not a reasonable pursuit: neuromorphic computing (dually analog/digital, plastic, stochastic, decentralized, and unobservable) is incompatible with the assumptions of conventional programming efforts (digital, immutable, noiseless, centralized, observable, see Figure \ref{fig:hero}).
% 2. Nothing ties everything together. TODO: expand
None of the methods and approaches we reviewed meet the criteria for a universally agreed-upon neuromorphic abstraction \cite{Mead_2023, SchumanEtAl2022Opportunities}.

% What we give the reader: conventional methods exist but they're not enough to cover all of 
However, decades of research in neuromorphic computing and engineering, provide important insights towards initial features of neuromorphic programming methods.
% go through the traces of each dimension
% none of them captures it, but we can combine them
Revisiting the requirements from Section \ref{s:motivation}, \textit{modern approaches must accommodate the simultaneously analog and digital nature of neural computation} to supplement digital models with differential equations in some shape or form.
This does not exclude digital instructions, but implies approximately continuous-time primitives, such as linearized time-stepped or variable-time models.
Second, \textit{the computational models must be able to capture the inherent change in the underlying substrate}.
This further challenges digital instructions because arbitrary bit-flips are detrimental to digital computation.
Online learning methods may help by continuously adapting to changes in the environment as well as changes in the underlying substrate.
% Approaches such as optimization or reservoir computing are more tractable.
%
Third, \textit{novel approaches should model noise on the signal-level and remain robust to small and unpredictable perturbations}.
This is challenging, but possible, to achieve in classical programs, and much more applied in machine learning and automated programming schemes, and even harnessed in probabilistic programming and stochastic computing. 
Fourth, \textit{any model should allow for event-based information processing}. Synchronicity in nature is extremely costly, but used almost everywhere in conventional programming paradigms except some reactive and concurrent languages.
Finally, \textit{the computation should depend solely on locally available information} rather than on global system states, as done by local plasticity rules used in neuromorphic hardware. 
Some machine learning and optimization methods, like reservoir computing or evolutionary optimization, also do not require a complete description of the computer's state.

% Machine learning and non-digital programming methods are applicable to \emph{analog} neuromorphic systems. 
% TODO: domain \& plasticity

%%% unify them?
To unite the physical dynamics and computational principles, fully neuromorphic abstractions are preferable, but probably not tractable before the ``hardware lottery'' has been won \cite{Hooker2020} and significant gains have been achieved with neuromorphic hardware compared against other computational substrates.
A more realistic scenario is that we can play on the strengths of multiple current approaches, and integrate them into one cohesive abstraction.
The neuromorphic system hierarchy \cite{ZhangEtAl2020} provides a common abstraction that can represent ANNs and SNNs, but is not designed for neurocomputational primitives, spiking neuromorphic architectures, or other neuromorphic abstractions.
The Neuromorphic Intermediate Representation \cite{pedersen_neuromorphic_2023} provides an abstraction for graph-structured continuous-time computations, which can support a wider variety of neuromorphic programming paradigms.
A key innovation in NIR is the integration of heterogeneous representations, such the Neural Engineering framework, Lava, Fugu, PyNN, and NeuroML.
The outcome is that multiple representations and paradigms can cooperate through a shared representation, flexible enough to cover multiple approaches, but unambiguous enough to only provide limits for discrepancies in the execution.
However, a complete/full neuromorphic abstraction must also offer support for computational graphs that change over time, which NIR presently does not offer.

A longstanding goal in computer science has been to program physical devices that mimic the efficiency and functionality of the brain \cite{Neumann1958,Jaeger2023}.
% It has long been the dream to program brain-like and efficient, physical devices.
While significant headway has been made towards instrumenting digital as well as non-digital systems, more work is needed to find robust programming methods for neuromorphics.
We hope that neuromorphic programmers can leverage the work outlined in this paper to build large-scale neuromorphic programs to tackle real-world tasks, and to further develop guiding principles and paradigms for neuromorphic programming. 

\section*{Acknowledgment}

S.A. thanks Herbert Jaeger, Guillaume Pourcel, and Mirko Goldmann for helpful comments and discussions.
S.A. gratefully acknowledges funding from the European Union's Horizon 2020 Research and Innovation Programme under the Marie Skłodowska-Curie grant agreement No. 860360 (POST DIGITAL).
J.P. would like to thank funding by the EC Horizon 2020 Framework Programme under Grant Agreements 785907 and 945539 (HBP) and NeuroPAC under the NSF grant
``AccelNet: Accelerating Research on Neuromorphic Perception, Action, and Cognition.''

% \bibliographystyle{acm/ACM-Reference-Format}
% \bibliography{cleaned_refs.bib}
\bibliographystyle{IEEEtran}
\bibliography{IEEEabrv,cleaned_refs}

% \begin{thebibliography}{00}
% \bibitem{b1} None
% \end{thebibliography}

\end{document}